\documentclass[10pt,twocolumn,letterpaper]{article}

\usepackage{iccv}
\usepackage{times}
\usepackage{epsfig}
\usepackage{graphicx}
\usepackage{amsmath}
\usepackage{amssymb}

\usepackage{graphicx}
\usepackage{amsmath}
\usepackage{amssymb}
\usepackage{caption}
\usepackage{booktabs}
\usepackage{amsfonts}       
\usepackage{nicefrac}       
\usepackage{microtype}      
\usepackage{graphicx}
\usepackage{diagbox}
\usepackage{algorithmic}
\usepackage{algorithm}
\usepackage{multicol}
\usepackage{enumerate}
\usepackage{times}
\usepackage{epsfig}
\usepackage{graphicx}
\usepackage{amsmath}
\usepackage{amssymb}
\usepackage{threeparttable}
\usepackage{enumitem}
\usepackage[section]{placeins}
\usepackage{multirow}
\usepackage{color}
\usepackage{array}
\usepackage{makecell}
\usepackage[table]{xcolor}
\usepackage{arydshln}
\usepackage[pagebackref=true,breaklinks=true,letterpaper=true,colorlinks,bookmarks=false]{hyperref}
\usepackage[capitalize]{cleveref}
\crefname{section}{Sec.}{Secs.}
\Crefname{section}{Section}{Sections}
\Crefname{table}{Table}{Tables}
\crefname{table}{Tab.}{Tabs.}

 \iccvfinalcopy 

\def\iccvPaperID{6265} 

\makeatletter
\def\hlinew#1{%
 \noalign{\ifnum0=`}\fi\hrule \@height #1 \futurelet
 \reserved@a\@xhline}
\makeatother%
\ificcvfinal\fi

\begin{document}

\title{From Sky to the Ground: A Large-scale Benchmark and Simple Baseline Towards Real Rain Removal}

\author{Yun Guo\textsuperscript{1,}\footnotemark[2], Xueyao Xiao\textsuperscript{1,}\footnotemark[2], Yi Chang\textsuperscript{1,}\footnotemark[1], Shumin Deng\textsuperscript{1}, Luxin Yan\textsuperscript{1}\\
\textsuperscript{1}National Key Laboratory of Science and Technology on Multispectral Information Processing,\\
School of Artificial Intelligence and Automation, Huazhong University of Science and Technology, China\\
{\tt\small \{guoyun, xiaoxueyao, yichang, shumindeng, yanluxin\}@hust.edu.cn}\\
\url{{https://github.com/yunguo224/LHP-Rain}}
}
\twocolumn[{%
\renewcommand\twocolumn[1][]{#1}%
\maketitle

\begin{center}
    \centering
    \captionsetup{type=figure}
    \includegraphics[width=1.0\textwidth]{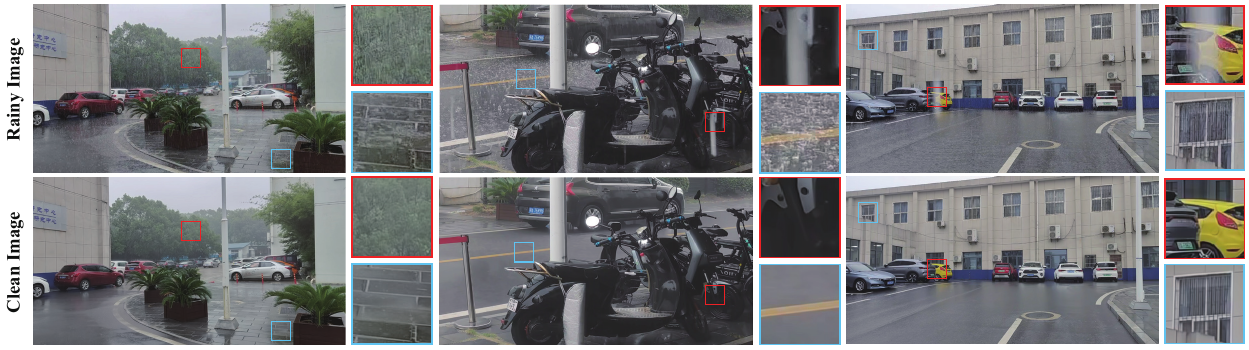}
    \captionof{figure}{Illustration of the proposed paired rainy/clean images dataset. The proposed real rain dataset is to cope with various rain categories, such as rain streaks, veiling effect, occlusion, and \textbf{ground splashing}. We recommend \textbf{zooming in} the figure on PC for better visualization.}
 \label{example}
\end{center}%
}]

\renewcommand{\thefootnote}{\fnsymbol{footnote}} 
\footnotetext[2]{These authors contributed equally to this work.} 
\footnotetext[1]{Corresponding author.} 

\begin{abstract}
\vspace{-0.4cm}
Learning-based image deraining methods have made great progress. However, the lack of large-scale high-quality paired training samples is the main bottleneck to hamper the real image deraining (RID). To address this dilemma and advance RID, we construct a Large-scale High-quality Paired real rain benchmark (LHP-Rain), including 3000 video sequences with 1 million high-resolution (1920*1080) frame pairs. The advantages of the proposed dataset over the existing ones are three-fold: rain with higher-diversity and larger-scale, image with higher-resolution and higher-quality ground-truth. Specifically, the real rains in LHP-Rain not only contain the classical rain streak/veiling/occlusion in the sky, but also the \textbf{splashing on the ground} overlooked by deraining community. Moreover, we propose a novel robust low-rank tensor recovery model to generate the GT with better separating the static background from the dynamic rain. In addition, we design a simple transformer-based single image deraining baseline, which simultaneously utilize the self-attention and cross-layer attention within the image and rain layer with discriminative feature representation. Extensive experiments verify the superiority of the proposed dataset and deraining method over state-of-the-art. 
\end{abstract}

\vspace{-0.4cm}
\section{Introduction}\label{sec1}
Single image deraining is to improve the imaging quality by separating rain from image background. In recent years, significant progress has been made in learning-based single image deraining by various sophisticated CNN architectures \cite{jiang2020multi, wang2020model, zamir2021multi, fu2021rain} and powerful Transformer models \cite{valanarasu2022transweather, xiao2022image}. Although these state-of-the-art supervised methods have achieved impressive results on simulated datasets, a fact cannot be ignored that those competitive methods perform unsatisfactory on diverse real rainy scenes. The core reason is the domain shift issue between the simplified synthetic rain and complex real rain \cite{ye2021closing, quan2021removing, wei2021deraincyclegan, ye2022unsupervised}.

\begin{table*}[t]
\footnotesize
  \renewcommand\arraystretch{1.1}
  \centering
  \setlength{\abovecaptionskip}{0pt}
  \setlength{\belowcaptionskip}{0pt}
  \caption{Summary of existing real rain datasets.}
  \begin{threeparttable}
    \setlength{\tabcolsep}{1.8mm}{
  \begin{tabular}{ccccccccc}
  \hlinew{1.5pt}
  Datasets & Year & Source &Sequence& Frame &Resolution &Rain Categories& Annotation & Paired\\
   \hlinew{1.5pt}
     RID/RIS\cite{li2019single} & 2019 &{Cam/Internet}&None& 4.5K &640*368&streak, raindrop &Object detection &- \\
      \cline{1-9}
     NR-IQA\cite{wu2020subjective} & 2020 &Internet &None& 0.2K&1000*680&streak, veiling&None &-\\
          \cline{1-9}
     Real3000\cite{liu2021unpaired} & 2021 &Internet &None&3.0K&942*654 &streak,  veiling& None &- \\
          \cline{1-9}
     FCRealRain\cite{ye2022unsupervised} & 2022 &Camera &None& 4.0K&4240*2400 &streak, veiling& Object detection & - \\
   \hlinew{1.5pt}
     SPA-Data\cite{wang2019spatial} & 2019 & {Cam/Internet} &170& 29.5K &256*256& streak& None&\checkmark \\
              \cline{1-9}
     RainDS\cite{quan2021removing} & 2021 & Cam &None&1.0K&1296*728&streak, raindrop&None&\checkmark \\
          \cline{1-9}
     GT-Rain\cite{ba2022gt-rain} & 2022 &Internet &202&31.5K&666*339& streak, veiling&None &\checkmark \\
          \cline{1-9}
     RealRain-1K\cite{li2022toward} & 2022 & {Cam/Internet}&1120& 1.1K &1512*973&streak, veiling, occlusion& None &\checkmark \\
          \cline{1-9}
             \hlinew{1.0pt}
     \textbf{LHP-Rain} & 2023 & Camera &\textbf{3000}&\textbf{1.0M} &1920*1080& \makecell[c]{streak, veiling, occlusion, \textbf{splashing}}& {Object detection/\textbf{Lane}}&\checkmark \\
   \hlinew{1.5pt}
  \end{tabular}}
  \end{threeparttable}
  \label{tab1}
  \vspace{-0.4cm}
 \end{table*}

To solve this problem, an intuitive idea is to try the best to make rain degradation model as real as possible \cite{garg2005does}. The researchers formulate the rain imaging procedure into a comprehensive rain simulation model \cite{fu2017removing, yang2017deep, hu2019depth, liu2018erase, li2019heavy}, in which different visual appearance of rain streaks\cite{fu2017removing}, accumulation veiling \cite{yang2017deep}, haze \cite{hu2019depth, li2019heavy}, and occlusion \cite{liu2018erase} factors are taken into consideration. Unfortunately, these linear simulation models still cannot well accommodate the distribution of realistic rains. For example, in Fig. \ref{example}, realistic rain streak is usually not exactly a regular line-pattern streak but possesses irregular non-uniform in terms of the intensity and width. Apart from the rain streaks, the existing rain simulation models could not handle the complicated rain splashing on the ground, which presents as dense point-shape texture, droplets or water waves, ruining visibility of traffic signs, such as lane lines, and also causes enormous negative effects for the high-level vision.

Another research line obtains the `clean' counterpart from the realistic rainy videos \cite{wang2019spatial, li2022toward}, which leverages the motion discrepancy between static image background and dynamic rain. Unfortunately, they simply employ naive filtering strategies such as percentile filter \cite{wang2019spatial} and median filter \cite{li2022toward}, resulting in unsatisfactory GT with residual rain or over-smoothing phenomenon. Moreover, the number and diversity of the existing real paired rain datasets are still limited. Few datasets have considered the rain splash on the ground, which is commonly observed in the real world but still rarely mentioned in deraining community. And the number of existing video sequences and image frames are not sufficient to cover diverse rains in terms of the varied rain angle, intensity, density, length, width and so on. Last but not least, the existing realistic rainy images are mostly downloaded from the Internet with low-quality: compression, watermark, low-resolution, without annotation and so on. As such, constructing a large-scale high-quality paired realistic rain dataset is highly necessary.

In this work, we construct a new large-scale high-quality paired real rain benchmark. The strength of our benchmark is threefold. First, the LHP-Rain contains diverse rain categories with very large-scale, including 3000 video sequences with over 1 million frame pairs. Second, apart from the conventional streak and veiling, our benchmark is capable of removing the representative challenging ground splashing rain in the real world. Third, the LHP-Rain is collected by the smartphone with high-resolution (1920*1080 pixels) and abundant objects under self-driving and surveillance scenes are captured for comprehensive evaluation. Moreover, we propose a novel robust low-rank tensor recovery method (RLRTR) for video deraining, which can generate higher-quality GT with better rain removal from sky to the ground and image structure preserving. We summary the main contributions as follows:

\begin{itemize}[leftmargin=10pt]
\item We construct a large-scale high-quality paired real rain benchmark for real single image deraining. To our best knowledge, LHP-Rain is the largest paired real rain dataset (3000 video sequences, 1 million frames) with high image resolution (1920*1080), and the first benchmark to claim and tackle the problem of ground splashing rain removal. 
\setlength{\itemsep}{-2pt}
\item We design a novel robust low-rank tensor recovery model for video deraining to better acquire paired GT. We provide detailed analysis to show RLRTR can better differ the rain from static background than previous datasets.
\setlength{\itemsep}{-2pt}
 \item We propose a new transformer-based single image deraining baseline, which exploits both self-attention and cross-layer attention between the rain and image layer for better representation. Extensive experiments on different real datasets verify the superiority of proposed method.
\end{itemize}

\section{Related Work}\label{sec2}
\begin{figure*}[htbp]
 \vspace{0cm}  
\setlength{\abovecaptionskip}{0.1 cm}   
\setlength{\belowcaptionskip}{-0.4 cm}   
  \centering
     \includegraphics[width=1.00\linewidth]{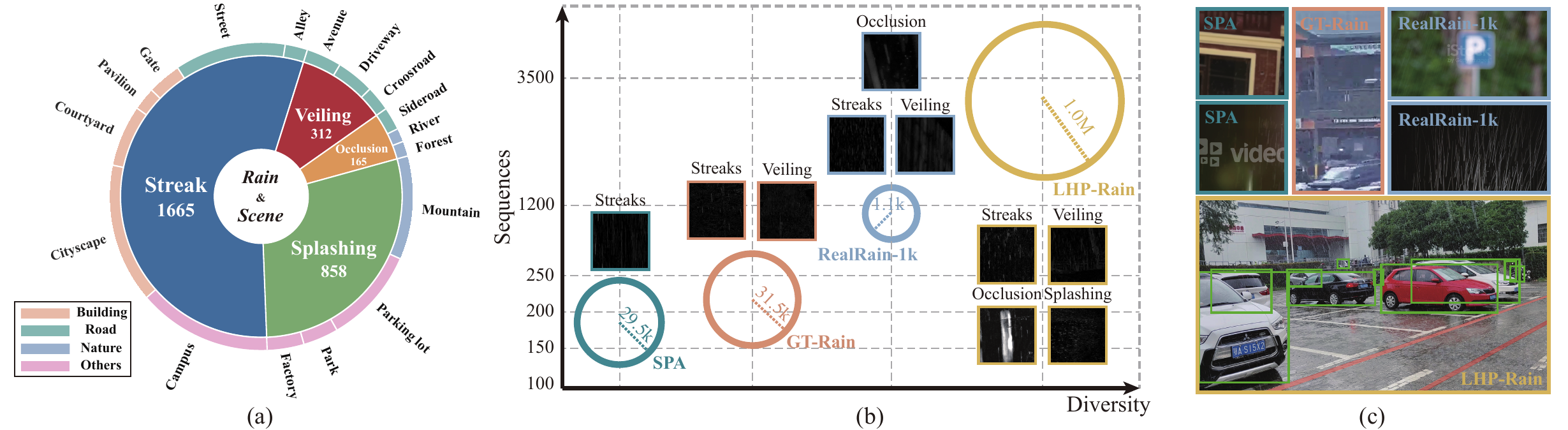}
  \caption{Features of the proposed benchmark LHP-Rain. (a) Distribution of rain and scene of the proposed benchmark. (b) Our proposed LHP-Rain outperforms others in terms of rain diversity and sequence amount.  (c) LHP-Rain collects high-resolution and annotated rainy images without copyright, compression and blur.}
  \label{DatasetComparison}
\end{figure*}

\textbf{Real rain datasets.}
At present, the researchers have mostly focused on the network architecture design, while relative fewer attention has been paid on the real rain dataset. The insufficiency of realistic rain dataset is the main bottleneck to hamper single image deraining. In Table \ref{tab1}, we provide a comprehensive summary of existing real rain datasets, which can be classified into two categories: rainy image only and paired rain-image. The former can be utilized via semi-supervised \cite{wei2019semi, huang2022memory} or unsupervised methods \cite{chen2022unpaired, ye2022unsupervised}. The latter can be conveniently utilized by the supervised training.

The key to paired real dataset is how to acquire the pseudo-`clean' image from its rainy counterpart. There are two main ways to construct the pairs: video-based generation (SPA-data \cite{wang2019spatial}, RealRain-1K \cite{li2022toward}), and time-interval acquisition (RainDS \cite{quan2021removing}, GT-Rain \cite{ba2022gt-rain}). All these datasets should ensure
that the camera is strictly immobile during the acquisition
process. SPA-data \cite{wang2019spatial} was the first presented paired real dataset which utilized the human-supervised percentile video filtering to obtain the GT. Instead of generation, GT-Rain \cite{ba2022gt-rain} collected the pairs of same scene under rainy and good weather, respectively. Similar idea has been adopt in RainDS \cite{quan2021removing} by manually mimicking rainfall with a sprinkler.

Despite the rapid development promoted by those pioneer datasets, there remain some important issues to be solved: insufficient number and rain diversity (not consider ground splashing). In this work, we contribute a large-scale high-quality paired real rain dataset (Section \ref{sec3.1}) with diverse rain and abundant objects. Moreover, a novel GT generation method is proposed with higher-quality pairs (Section \ref{sec3.3}).

 \setlength{\parskip}{0em}
\textbf{Single image deraining.}
The single image deraining methods have made great progress in last decade including the optimization models \cite{luo2015removing, li2016rain, chang2017transformed}, deep convolutional network \cite{hu2019depth, wang2020model, fu2021rain} and transformer \cite{valanarasu2022transweather, xiao2022image}. Fu \emph{et al}. \cite{fu2017removing} first introduced the residual deep CNN for single image deraining. Latter, the researchers have further improved the network depth by stacking similar modules, such as the well-known recurrent \cite{li2018recurrent, yang2019single} and multi-stage progressive \cite{ren2019progressive, zamir2021multi} strategies. Meanwhile, the multi-scale has been widely explored to improve the representation such as the multi-scale fusion \cite{jiang2020multi} and wavelet decomposition \cite{yang2019scale}. Further, the side information about the rain attribute: density \cite{zhang2018density}, depth \cite{hu2019depth}, directionality \cite{wang2020rain}, location \cite{yang2019joint}, non-local \cite{li2018non} have been extensively utilized to enhance the deraining performance.

\setlength{\parskip}{0em}
Benefiting from self-attention mechanise for long-range relationships modelling, transformer-based methods have achieved significant performance for single image deraining \cite{wang2022uformer, valanarasu2022transweather, xiao2022image, chen2023DRSformer}. Very recently, Xiao \emph{et al}.\cite{xiao2022image} proposed an image deraining transformer (IDT) with relative position enhanced and spatial-based multihead self-attention. Chen \emph{et al}. \cite{chen2023DRSformer} proposed a sparse Transformer architecture to solve the redundant feature issue.  In this work, we propose a simple yet effective dual-branch transformer baseline which simultaneously utilizes the self-attention within rain/image layer and cross-layer attention between the rain and image layer, so as to jointly improve the discriminative disentanglement between the rain and image layer (Section \ref{sec4}).

\setlength{\parskip}{0em}
\section{Large-scale high-quality paired Benchmark}
\subsection{Benchmark Collection and Statistics}\label{sec3.1}
\begin{figure*}[t]
 \vspace{0cm}  
\setlength{\abovecaptionskip}{0 cm}   
\setlength{\belowcaptionskip}{-0.4 cm}   
  \centering
     \includegraphics[width=1.00\linewidth]{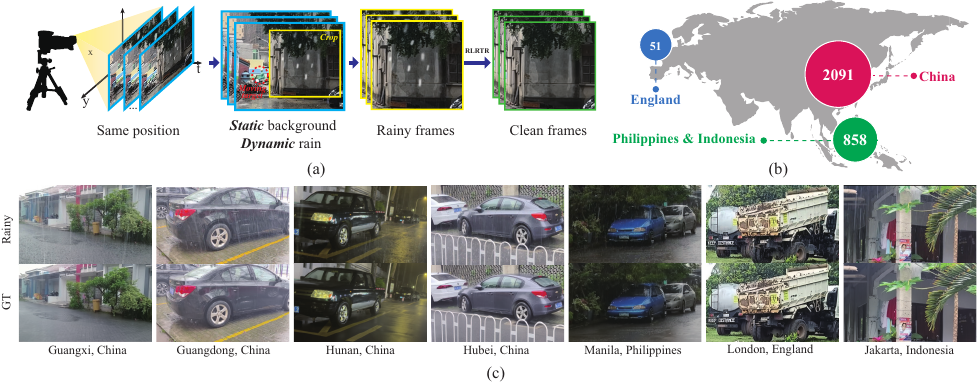}
  \caption{Illustration of the proposed benchmark LHP-Rain. (a) Overall procedure of obtaining rainy/clean image pair. (b) Quantity and location distribution of LHP-Rain.  (c) Rainy/GT samples of LHP-Rain from different locations. Scenes are varied from nature to city, over day and night with diverse rain patterns from sky to the ground. }
  \label{Collection}
\end{figure*}

Due to the difficulty and inconvenience of collecting real rain videos, the video sequences and frames of existing paired real rain datasets are still limited as shown in Table \ref{tab1}. In this work, we collect the real rain sequences by smartphones with 24mm lens focal length, sampled in 30 fps. The data collection process is illustrated in Fig. \ref{Collection}(a). Firstly, to keep the camera immobile, we employ tripod to capture real rain videos with static background (no moving object except rain). For each sequence, we record approximate 15 seconds and extract the intermediate steady 10s into our dataset. Then, we manually pick out moving object to remove unexpected disturbance. Finally, we employ the proposed RLRTR (Section \ref{sec3.3}) to obtain high-quality GT.

Overall, we collect 3000 video sequences with approximate 1 million frames across 8 cities from 4 countries around the world, China (2091 sequences), England (51 sequences), Philippines and Indonesia (858 sequences). We visualize the per-country image counts and location distribution of LHP-Rain in Fig. \ref{Collection}(b). The rainfall levels are varied from light rain (10mm/day) to rainstorm (300mm/day) due to diversity of local climates. Over 17 typical scenes are captured, including the parking lot, street, alley, playground, courtyard, and forest, etc. Besides, 2490 sequences are captured at daytime and 510 sequences are at night. More Rainy/GT pairs from LHP-Rain are displayed in Fig. \ref{Collection}(c). With the changing of location, backgrounds are varied from nature to city with diverse rain patterns, such as rain streak, veiling effect, occlusion and splashing. Note that, although the background is the same for each video sequence, the rain in each frame is vastly different from the appearance, including streak, veiling, occlusion and splashing. Here we separate 2100 sequences as training set, 600 sequences as validation set and the other 300 sequences as test set. To further visualize the quantity distribution of rain and scene for each sequence, a sunburst chart is illustrated in Fig. \ref{DatasetComparison}(a).  
\subsection{Benchmark Features}\label{sec3.2}
\noindent\textbf{Rain with higher-diversity and larger-scale.} We concern not only the sequence number and total frames but also the rain diversity of realistic rain, which both have great impact on the generalization for real-world rain. In Fig. \ref{DatasetComparison}(b), we show the statistic distribution of rain diversity and sequences/frames in typical real paired rain dataset. All datasets concern about the noticeable rain streak, especially SPA-data \cite{wang2019spatial}. RainDS \cite{quan2021removing} additionally takes the raindrop into consideration with 1000 frames, while GT-Rain \cite{ba2022gt-rain} and RealRain-1K \cite{li2022toward} further capture the accumulation veiling artifact in heavy rain. LHP-Rain contains not only rain streak and veiling in the sky, but also challenging highlight occlusion and splashing water on the ground. To our best knowledge,  the proposed LHP-Rain is the first benchmark to collect and tackle ground splashing rain in the real world, which is commonly ignored by previous existing datasets. 

\noindent\textbf{Image with higher-resolution and abundant objects.}
The existing datasets pay much attention to the rain, ignoring that the high-quality image is also what we really need. Unfortunately, existing realistic rainy images are generally downloaded from the Internet with various problems about the image: compression artifact, watermark, low-resolution, out-of-focus blur, without objects to name a few, which may cause challenges for high-level vision applications. In Fig. \ref{DatasetComparison}(c), we show the typical examples in each dataset. The image background of RealRain-1K \cite{li2022toward} suffers from serious out-of-focus blur, since they have manually focused on the rain on purpose. GT-Rain \cite{ba2022gt-rain} contains obvious compression artifact, since it origins from the compressed Youtube stream. There are numerous scenes with narrow views and watermark in the SPA-data \cite{wang2019spatial}, because they only release patches (256*256) cropped from original frames.

To improve the image quality, we personally capture high-resolution (1920*1080) realistic rain videos by smartphones. Moreover, LHP-Rain is not only designed for rain restoration, but also important for object detection and segmentation tasks under adverse weather, with abundant objects which are oriented for self-driving and video surveillance scenes. Thus, we provide annotations for object detection and lane segmentation. Five typical objects including person, car, bicycle, motorcycle and bus are annotated by bounding box with 326,961 instances totally. For lane segmentation, we annotate 24,464 lane masks to evaluate the effect of rain splashing removal. Note that the same object in different frames will be regarded as different instances because rain is inconstant and changing frame by frame. 

 \begin{figure}[t]
 \vspace{0cm}  
\setlength{\abovecaptionskip}{0.1 cm}   
\setlength{\belowcaptionskip}{-0.6 cm}   
  \centering
     \includegraphics[width=1.00\linewidth]{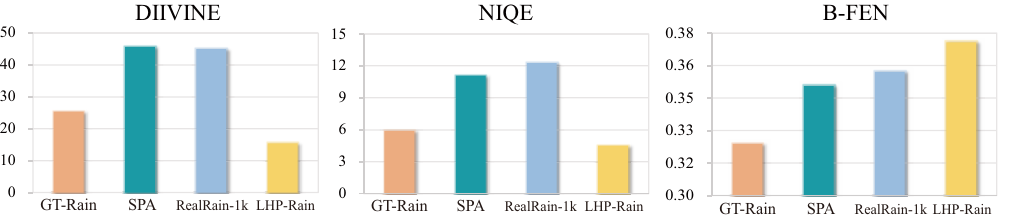}
  \caption{The GT quality of LHP-Rain is superior to others on DIIVINE (lowest), NIQE (lowest) and B-FEN (highest).}
  \label{Statistic}
\end{figure}

\begin{figure*}[htbp]
  \centering
     \includegraphics[width=1.00\linewidth]{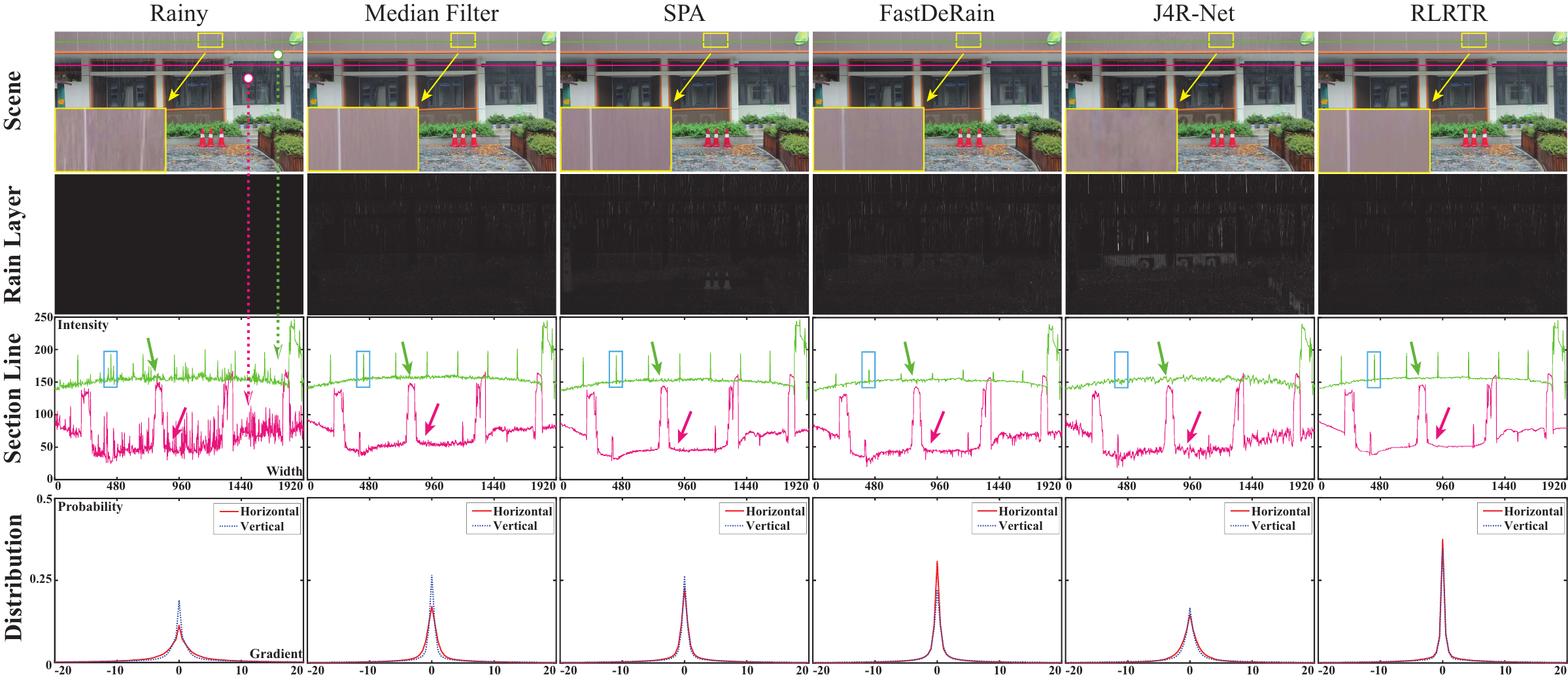}
  \caption{Analysis of different video deraining results on our dataset. From left to right, the first column is the original rainy frame, and the remaining five columns represent different methods, namely Median Filter, SPA, FastDeRain, J4R-Net and the proposed RLRTR. From top to bottom, the first row shows the deraining results, the second row is the rain layer of the deraining results, the third row denotes the section line of the deraining results and the last row represents the horizontal and vertical gradient distributions of the deraining results.}
  \label{ComparisionGT}
  \vspace{-0.4cm}
\end{figure*}

\noindent\textbf{Higher-quality ground-truth.}
The quality of GT is critical for paired real rain dataset. It is difficult to determine what is good or bad GT in an absolutely fair way. In this paper, we assume that the better the rain removal, the better the image quality is. Therefore, we employ several no reference image quality assessments: DIIVINE \cite{moorthy2011blind}, NIQE \cite{mittal2012making} and B-FEN \cite{wu2020subjective}, to evaluate the image quality of the rain-free image. The former two are hand-crafted based general image quality indexes, and the last one B-FEN is the learning based index especially designed for de-raining quality assessment. We select all the video backgrounds in SPA-data \cite{wang2019spatial}, GT-Rain \cite{ba2022gt-rain}, RealRain-1K \cite{li2022toward} and LHP-Rain for evaluation. In Fig. \ref{Statistic}, we can observe that the proposed LHP-Rain consistently obtains the best results in terms of different evaluation indexes which strongly support the higher-quality of the GT.

\subsection{Robust Low-rank Tensor Recovery Model}\label{sec3.3}

Given the rainy video $\boldsymbol{\mathcal{O}}\in{{\mathbb{R}}^{h \times w \times t}}$, the key is how to properly obtain the paired GT. The existing methods simply employ the naive filtering technique benefiting from the temporal consistency of the static background. Due to the slight camera vibration caused by the wild wind, we further leverage an affine transformation operator $\tau$ \cite{yong2017robust} to achieve the pixel-level alignment of each frame. Thus, a multi-frame rainy video can be described as the following formula:
 \begin{table*}[htbp]
\renewcommand\arraystretch{1.25}
\footnotesize
  \centering
  \setlength{\abovecaptionskip}{0pt}
  \setlength{\belowcaptionskip}{0pt}
  \caption{Quantitative comparisons with SOTA supervised methods on paired real datasets SPA-data (A), GT-Rain (B) and proposed LHP-Rain (C) under 9 different task settings. X$\rightarrow$Y means training on the dataset X and testing on the dataset Y. The degraded results of the three datasets are also provided. Top $1_{st}$ and $2_{nd}$ results are marked in \textcolor{red}{\textbf{red}} and \textcolor{blue}{\textbf{blue}} respectively.}
  \begin{threeparttable}
    \setlength{\tabcolsep}{0.55mm}{
  \begin{tabular}{ccccccccccccccccccc}
   \hlinew{1.5pt}
   \multirow{2}{*}{Method} & \multicolumn{2}{c}{A$\rightarrow$A} & \multicolumn{2}{c}{B$\rightarrow$A} & \multicolumn{2}{c}{C$\rightarrow$A} & \multicolumn{2}{c}{A$\rightarrow$B}& \multicolumn{2}{c}{B$\rightarrow$B} & \multicolumn{2}{c}{C$\rightarrow$B}& \multicolumn{2}{c}{A$\rightarrow$C} & \multicolumn{2}{c}{B$\rightarrow$C}& \multicolumn{2}{c}{C$\rightarrow$C} \\
  \cline{2-19}
      & PSNR & SSIM & PSNR & SSIM & PSNR & SSIM & PSNR & SSIM& PSNR & SSIM & PSNR & SSIM& PSNR & SSIM& PSNR & SSIM& PSNR & SSIM\\
   \hlinew{1.5pt}
Rainy Image &\multicolumn{6}{c}{32.60 / 0.9173}&\multicolumn{6}{c}{19.48 / 0.5849}&\multicolumn{6}{c}{29.97 / 0.8497}\\
  \cline{1-19}
    SPANet & 38.53 &  0.9875 & 22.93 & 0.8207&31.46  &0.9612  & 20.01 & 0.6148&21.51&0.7145&19.20&0.5706&28.00&0.8905&20.10&0.8061&
    31.19&\textcolor{blue}{\textbf{0.9346}}\\
    \rowcolor{gray!25}
    PReNet& 37.05 &  0.9696 & 22.44 &0.7713 &\textcolor{blue}{\textbf{32.46}}&0.9387&20.29  & 0.5860 &20.65&0.6005&19.34&0.5530&27.57&0.8595&20.91&0.7222&32.13&0.9177\\
    RCDNet& 39.74 &  0.9661 &22.51 & 0.8392& 32.30 & 0.9378 & 20.09 &0.5785 &21.04&0.6106&19.09&0.5264&25.46&0.7959&21.38&0.8047&32.34&0.9152\\
   \rowcolor{gray!25}
    JORDER-E & 40.63 & 0.9794  & 23.47 & 0.7426& 31.23 &0.9234  & 19.98 & 0.5799&21.24&0.6854&18.76&0.4861&27.13&0.8531&22.14&0.8433&31.24&0.8847\\
    MPRNet&46.06  & 0.9894  &24.27 &0.8428 &  32.37& 0.9379 &  19.87&0.6286 &22.00&0.6515&19.47&0.5889&28.41&0.8807&\textcolor{blue}{\textbf{23.82}}&0.8052&33.34&0.9309\\
   \rowcolor{gray!25}
    GT-Rain & 37.21 &  0.9827 & \textcolor{blue}{\textbf{25.30}} & \textcolor{red}{\textbf{0.9243}}& 26.46 & 0.9145 & 20.07 &\textcolor{blue}{\textbf{0.6941}}&\textcolor{blue}{\textbf{22.51}}&\textcolor{blue}{\textbf{0.7300}}&\textcolor{blue}{\textbf{21.14}}&0.5698&28.62&0.8675&23.19&0.8098&32.18&0.9132\\
    Uformer-B &\textcolor{blue}{\textbf{46.42}} &\textcolor{blue}{\textbf{0.9917}} & 24.08& 0.8979& 32.21 &\textcolor{blue}{\textbf{0.9667}}  &19.70  &0.6875 &21.60&0.7124&19.10&\textcolor{red}{\textbf{0.6622}}&\textcolor{blue}{\textbf{28.74}}&\textcolor{red}{\textbf{0.9262}}&22.91&\textcolor{red}{\textbf{0.8734}}&\textcolor{blue}{\textbf{33.56}}&0.9317\\
    \rowcolor{gray!25}
    IDT& 45.74&0.9889 &23.80 &0.8334&32.38  & 0.9422 &\textcolor{blue}{\textbf{20.34}} &0.6306 &21.98&0.6536&19.44&0.5977&26.90&0.8742&23.34&0.7897&33.02&0.9310\\
    \cline{1-19}
    \textbf{SCD-Former} &\textcolor{red}{\textbf{46.89}}&\textcolor{red}{\textbf{0.9941}}&\textcolor{red}{\textbf{26.13}}&\textcolor{blue}{\textbf{0.9122}}&\textcolor{red}{\textbf{34.38}}  &\textcolor{red}{\textbf{0.9798}}  &\textcolor{red}{\textbf{20.98}}&\textcolor{red}{\textbf{0.6985} }&\textcolor{red}{\textbf{22.79}}&\textcolor{red}{\textbf{0.7684}}&\textcolor{red}{\textbf{21.71}}&\textcolor{red}{\textbf{0.6893}}&\textcolor{red}{\textbf{29.41}}&\textcolor{blue}{\textbf{0.9127}}&\textcolor{red}{\textbf{23.56}}&\textcolor{blue}{\textbf{0.8626}}&\textcolor{red}{\textbf{34.33}}&\textcolor{red}{\textbf{0.9468}}\\
       \hlinew{1.5pt}
  \end{tabular}}
  \end{threeparttable}
  \label{tab2}
  \vspace{-0.4cm}
 \end{table*}
 
   \vspace{-0.3cm}
\begin{equation}
  \vspace{-0.2cm}
\setlength{\abovedisplayskip}{0pt}
\setlength{\belowdisplayskip}{0pt}
{\mathcal{O}} \circ \tau  = {\mathcal{B}} + {\mathcal{R}} + {\mathcal{N}},
\label{eq:syn3}
\end{equation}

where ${\mathcal{B}}\in{{\mathbb{R}}^{h \times w \times t}}$ is the rain-free video, ${\mathcal{R}}\in{{\mathbb{R}}^{h \times w \times t}}$ represents the rains, ${\mathcal{N}}\in{{\mathbb{R}}^{h \times w \times t}}$ denotes the random noise, and $\tau$ denotes the affine transformation to ensure the rainy video of each frame is pixel-level aligned. In this work, we formulate the video deraining into inverse problem via the \emph{maximum-a-posterior} as follow:
\begin{equation}
\setlength{\abovedisplayskip}{2pt}
\setlength{\belowdisplayskip}{2pt}
 \mathop {\min }\limits_{{\cal B},{\cal R}, \tau} \frac{1}{2}||{\cal B} + {\cal R} - {\cal O \circ \tau}|| _F^2 +\omega P_b ({\cal B}) +\mu P_r ({\cal R}),
\label{eq:syn4}
\end{equation}
where $P_b$ and $P_r$ are the prior knowledge for the image and rain, respectively, $\omega$ and $\mu$ are the corresponding hyper-parameters. As for the aligned rainy video, when there are no moving objects except the rain, the rain-free background image is the same for all rainy frames. That is to say, clean video $\cal{B}$ has extreme \emph{global} low-rank property along the temporal dimension, ideally its rank is equal to one for each scene. On the other hand, the clean video $\cal {B}$ also has very \emph{non-local} low-rank property along the spatial dimension, due to the self-similarity widely employed in image restoration \cite{dabov2007image}. Moreover, we further take the \emph{local} smoothness of the video $\cal {B}$ into consideration via the total variation (TV) regularization \cite{chang2017transformed}. Thus, the joint global-nonlocal-local prior along both the spatial and temporal dimension has been fully exploited for better representation of the static video $\cal {B}$:
\begin{equation}
\setlength{\abovedisplayskip}{6pt}
\setlength{\belowdisplayskip}{6pt}
\begin{aligned}
&\resizebox{0.865\hsize}{!}{$P_b ({\cal B}) =  \omega \mathop \sum \limits_i \left(\frac{1}{{\lambda _i^2}}||{{\cal S}_i}{\cal B}{ \times _3}Q_i - {{\cal J}_i}|| _F^2 +||{\cal J}_i||_{tnn}\right) + \gamma ||{\nabla_t}{\cal B}||_1,$}
\end{aligned}
\label{eq:syn4-1}
\end{equation}
where ${{\cal S}_i}{\cal B} \in {{\mathbb{R}}^{{p^2} \times k \times t}}$ is the constructed 3-D tensor via the non-local clustering of a sub-cubic ${u_i} \in {{\mathbb{R}}^{p \times p \times t}}$ \cite{chang2017hyper}, $p$ and $k$ are the spatial size and number of the sub-cubic respectively, $Q_i \in {{\mathbb{R}}^{d \times t}}(d \ll t)$ is an orthogonal subspace projection matrix used to capture the temporal low-rank property, $\times _3$ is the tensor product along the temporal dimension \cite{kolda2009tensor}, ${{{\cal J}_i}}$ represents the low-rank approximation variable, $||\bullet||_{tnn}$ means the tensor nuclear norm for simplicity \cite{chang2017hyper}, ${\nabla _t}$ is the difference operator, $\gamma$ and $\lambda _i$ is the regularization parameters. As for the rain $\cal{R}$, we formulate it as the sparse error \cite{wright2009robust} via the $L_1$ sparsity. Thus, the Eq. (\ref{eq:syn4}) can be expressed as:
\begin{equation}
\setlength{\abovedisplayskip}{6pt}
\setlength{\belowdisplayskip}{6pt}
\begin{aligned}
&\resizebox{0.85\hsize}{!}{$\left\{ {\hat {\cal B},\hat {\cal R},{{\hat {\cal J}}_i},\hat{\tau}, \hat Q_i} \right\} = \arg \mathop {\min }\limits_{{\cal B},{\cal R},{{\cal J}_i},\tau,Q_i} \frac{1}{2}|| {\cal B} + {\cal R} - {\cal O \circ \tau}|| _F^2$}\\
&\resizebox{0.865\hsize}{!}{$ + \mu ||{\cal R}{|| _1} + \omega \mathop \sum \limits_i \left( {\frac{1}{{\lambda _i^2}}||{{\cal S}_i}{\cal B}{ \times _3}Q_i - {{\cal J}_i}||_F^2 + ||{\cal J}_i||_{tnn}} \right) + \gamma ||{\nabla_t}{\cal B}{|| _1}.$}
\end{aligned}
\label{eq:syn4-2}
\end{equation}
\noindent\textbf{Optimization.}
Due to the difficulty of estimating multiple variables directly, we adopt the alternating minimization scheme to solve the Eq. (\ref{eq:syn4-2}) with respect to each variable.

1) \emph{Affine Transformation} $\tau$: Since ${\cal O \circ \tau}$ is a nonlinear geometric transform, it's difficult to directly optimize $\tau$. A common technique is to linearize around the current estimate and iterate as follows: ${\mathcal{O}} \circ \tau  + \nabla\mathcal{O} \triangle \tau= {\mathcal{B}} + {\mathcal{R}} + {\mathcal{N}}$ \cite{peng2012rasl}, where $\nabla\mathcal{O}$ is the Jacobian of the image $\mathcal{O}$ with respect to $\tau$. This method iteratively approximates the original nonlinear transformation with a locally linear approximation \cite{peng2012rasl}.

2) \emph{Rain Estimation} $\mathcal{R}$: By ignoring variables independent of $\mathcal{R}$, we can obtain following subproblem:
\begin{equation}
\setlength{\abovedisplayskip}{6pt}
\setlength{\belowdisplayskip}{6pt}
 \resizebox{0.85\hsize}{!}{$\hat {\cal R}  = \arg \mathop {\min }\limits_{{\cal R}} \frac{1}{2}|| {\cal B} + {\cal R} - {\cal O \circ \tau}|| _F^2 + \mu ||{\cal R}{|| _1}.$}
\label{eq:syn5-1}
\end{equation}
Eq. (\ref{eq:syn5-1}) is a $L_1$ minimization problem which can be easily solved by soft thresholding with closed-form solution \cite{lin2011linearized}.
 
3) \emph{Subspace Projection} $Q_i$: We enforce the orthogonal constraint on $Q_i^TQ_i=I$ with the following subproblem:
\begin{equation}
\setlength{\abovedisplayskip}{6pt}
\setlength{\belowdisplayskip}{6pt}
{\hat Q_i}= \arg \mathop {\min }\limits_{Q_i^TQ_i=I} \frac{1}{{\lambda _i^2}}||{{\cal S}_i}{\cal B}{ \times _3}Q_i - {{\cal J}_i}||_F^2.
\label{eq:syn5-2}
\end{equation}
According to \cite{xie2017kronecker}, Eq. (\ref{eq:syn5-2}) has the closed-form solution, which can be obtained by the \emph{rank-d} singular value decomposition of the folding matrix of ${{{\cal S}_i}{\cal B}}$, where \emph{d} is the measurement of the intrinsic subspace of the temporal dimension. In this work, we empirically set $d \leq 3$.

4) \emph{Low-rank Approximation} ${{\cal J}_i}$: Dropping the irrelevant variables, we can get following subproblem:
\begin{equation}
\setlength{\abovedisplayskip}{6pt}
\setlength{\belowdisplayskip}{6pt}
 \resizebox{0.865\hsize}{!}{${{\hat {\cal J}}_i} = \arg \mathop {\min }\limits_{{{\cal J}_i}} \frac{1}{{\lambda _i^2}}\parallel {{\cal S}_i}{\cal B}{ \times _3}Q_i - {{\cal J}_i}\parallel _F^2 + ||{\cal J}_i||_{tnn}.$}
\label{eq:syn5-3}
\end{equation}
This is a typical tensor nuclear norm minimization problem, can be solved by singular value thresholding algorithm \cite{cai2010singular, chang2017hyper}.

5) \emph{Clean Video Estimation} ${{\cal B}}$: We fix the other variables and optimize ${{\cal B}}$ with the following subproblem:
\begin{equation}
\setlength{\abovedisplayskip}{6pt}
\setlength{\belowdisplayskip}{6pt}
\resizebox{0.865\hsize}{!}{$  \mathop {\min }\limits_{{\cal B}} \frac{1}{2}|| {\cal B} + {\cal R} - {\cal O \circ \tau}|| _F^2  + \omega \mathop \sum \limits_i {\frac{1}{{\lambda _i^2}}||{{\cal S}_i}{\cal B}{ \times _3}Q_i - {{\cal J}_i}||_F^2}  + \gamma ||{\nabla_t}{\cal B}{|| _1}.$}
\label{eq:syn5-4}
\end{equation}
Due to the non-differentiability of the $L_1$ norm in Eq. (\ref{eq:syn5-4}), we apply the ADMM \cite{lin2011linearized} to decouple this problem into several sub-problems with closed-form solutions. Please refer to the supplementary material for the whole algorithm details.
 
\noindent\textbf{Discussion.} Figure \ref{ComparisionGT} illustrates the comparison results of representative video deraining methods: filter-based methods (median filter\cite{li2022toward}, SPA\cite{wang2019spatial}), optimization-based (FastDeRain\cite{jiang2018fastderain}) and learning-based methods (J4R-Net\cite{liu2018erase}). The first and second rows show the video deraining of the image and rain layer, respectively.  RLRTR removes almost all the rain from sky to the ground, including rain streaks and splashing, and preserve the image details well, while other methods more or less have rain streaks residual but also cause noticeable damage to the image. In third row, we randomly choose two 1D section lines of the deraining results. The section line of RLRTR is smoother with less burr than other methods. Moreover, the SPA, FastDeRain and the J4R-Net unexpectedly attenuate the spike signal of the share edge, while the proposed RLRTR has well preserve the spike signal. It is well-known the natural image is isotropic and its gradient distribution along different directions should be close to each other \cite{torralba2003statistics}. Compared with other results, in forth row the gradient distributions along the vertical and horizontal directions of RLRTR are most similar to each other, which further indirectly verify the naturalness of the deraining result and produce better paired clean-rainy GT. 
 
   \begin{figure}[t]
  \vspace{0cm}  
\setlength{\abovecaptionskip}{0 cm}   
\setlength{\belowcaptionskip}{-0.3 cm}   
  \centering
     \includegraphics[width=1.00\linewidth]{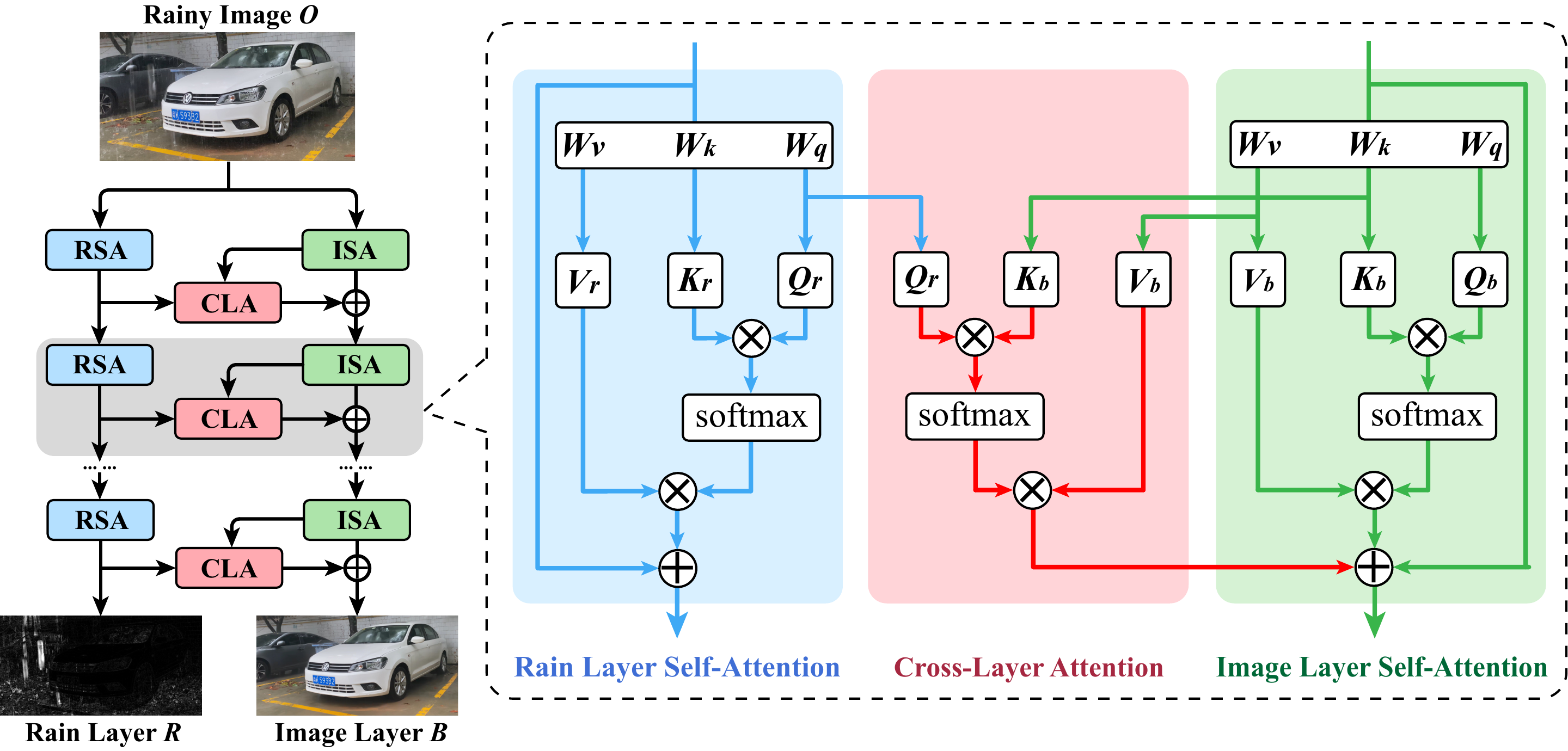}
  \caption{Overall framework of the SCD-Former. It utilizes self and cross-layer attention from rain layer to image layer which serves as side information to recover image layer. }
  \label{network}
\end{figure}

 \begin{figure*}[t]
  \vspace{0cm}  
\setlength{\abovecaptionskip}{0 cm}   
\setlength{\belowcaptionskip}{-0.4 cm}   
  \centering
     \includegraphics[width=1.00\linewidth]{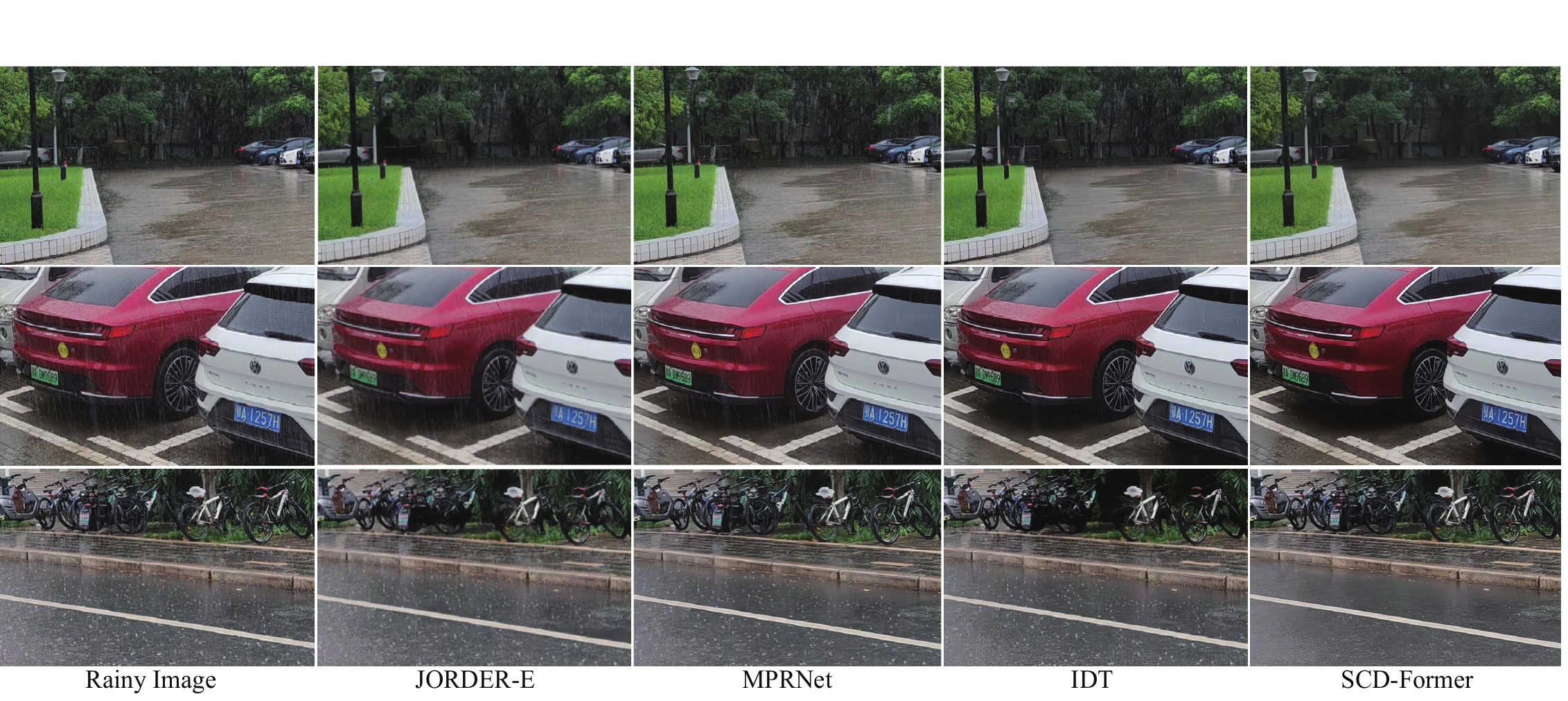}
  \caption{Visual comparisons on LHP-Rain. Comparing with state-of-the-arts, SCD-Former achieves more visual pleasing deraining results and it is capable of removing the highlight occlusion on the car and the splashing water on the ground.}
  \label{RealRain}
\end{figure*}

\section{SCD-Former: Image Deraining Baseline}\label{sec4}
The degradation procedure can be formulated as:
\begin{equation}
\setlength{\abovedisplayskip}{2pt}
\setlength{\belowdisplayskip}{2pt}
\emph{O}= \emph{B} + \emph{R}.
\label{eq:rain}
\end{equation}

It has been proved that the rain such as location \cite{yang2019joint} serving as an attention would be informative in CNN-based image restoration. In this work, we show the rain attention would also be beneficial in transformer. In Fig. \ref{network}, we design a simple two-stream Self- and Cross-attention Deraining Transformer (SCD-Former), in which the two-stream network is designed to restore rain and image layer respectively. On one hand, we utilize the self-attention in each rain/image stream independently; on the other hand, we further exploit the cross-layer attention between the rain and image streams. Thus, the rain collaboratively interactive with the image layer to further improve the discriminative representation.

\noindent\textbf{Self-attention and cross-layer attention.} 
In this work, we exploit self-attention on rain and image layer as Rain layer Self-Attention (RSA) and Image layer Self-Attention (ISA). Given the input feature \emph{X}, it will be projected into query (\emph{Q}), key (\emph{K}) and value (\emph{V}) by three learnable weight matrices ${W_q}$, ${W_k}$ and ${W_v}$. Then dot-product, scaling and softmax among \emph{Q}, \emph{K} and \emph{V} will be conducted. The self-attention function is defined as:
\begin{equation}
\setlength{\abovedisplayskip}{0pt}
\setlength{\belowdisplayskip}{2pt}
\emph{Attention}(\emph{Q, K, V})= \emph{softmax}(\frac{\emph{Q}\emph{K}^{\emph{T}}}{\sqrt{\emph{d}_{\emph{k}}}})\emph{V}.
\label{eq:SA}
\end{equation}
We further design a Cross-Layer Attention (CLA) module which bridges the attention relationship between rain and image layer. The CLA conducts attention operation among {$\emph{Q}_{\emph{r}}$} from rain layer and {$\emph{K}_{\emph{b}}$}, {$\emph{V}_{\emph{b}}$} from image layer as follow:
\vspace{-0.2cm}
\begin{gather}
\setlength{\abovedisplayskip}{0pt}
\setlength{\belowdisplayskip}{0pt}
\emph{CLA}(\emph{Q}_{\emph{r}}, \emph{K}_{\emph{b}}, \emph{V}_{\emph{b}}) = \emph{softmax}(\frac{{\emph{Q}_{\emph{r}}}{\emph{K}_{\emph{b}}}^{\emph{T}}}{\sqrt{\emph{d}_{\emph{k}}}})\emph{V}_{\emph{b}}.
\label{eq:CLA}
\end{gather}
The token of rain layer serves as a query token {$\emph{Q}_{\emph{r}}$} to interact with the patch tokens {$\emph{K}_{\emph{b}}$} and {$\emph{V}_{\emph{b}}$} from the image layer through attention mechanism. By calculating the correlation degree between both layer, the highly attentive location of rain residual can be acquired, which provides an extra prior for enhanced feature representation. Note that, the CLA module has been stacked over the whole network. Compared with previous work, SCD-Former exploits not only the self-attention but also cross-layer attention of the rain and image layer for better restoration. 

\setlength{\parskip}{0em}
\noindent\textbf{Implementation details.} We train the network using the Charbonnier loss\cite{charbonnier1994two} supervised by the ground truth of rain and image layer:
\begin{equation}
\setlength{\abovedisplayskip}{2pt}
\setlength{\belowdisplayskip}{2pt}
\mathcal{L} = ||O - B - R||_F^2+ \lambda_{r}||R - \hat{R}||_1+\lambda_{b}||B - \hat{B}||_1.
\label{eq:loss2}
\end{equation}
The framework is implemented with two RTX 3090 GPUs. We set the hyperparameter $\lambda_{b}$ and $\lambda_{r}$ as 1. The images are randomly cropped into 256 * 256 for training. The learning rate of network is set as 0.0002. The Adam optimizer is adopted for optimization with a batch size of 32.
\vspace{-0.1cm}
\section{Experiments}\label{sec5}
\vspace{-0.1cm}
 \begin{figure*}[t]
 \vspace{0cm}  
\setlength{\abovecaptionskip}{0.1 cm}   
\setlength{\belowcaptionskip}{-0.5 cm}   
  \centering
     \includegraphics[width=1.00\linewidth]{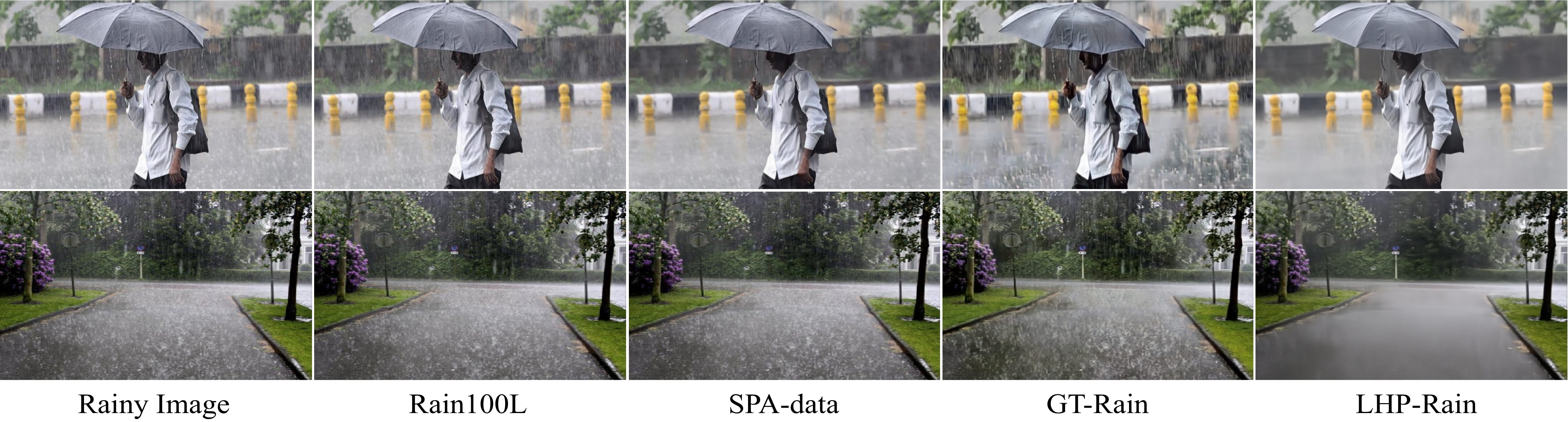}
  \caption{Evaluation of the diversity of the LHP-Rain. We train SCD-Former on different datasets: Rain100L, SPA-data, GT-Rain and LHP-Rain, and test on other datasets. The model trained on LHP-Rain has achieved better deraining results. }
  \label{DatasetGeneralization}
\end{figure*}

\noindent\textbf{Datasets.} We conduct the experiments on paired datasets SPA-data\cite{wang2019spatial}, GT-Rain\cite{ba2022gt-rain} and LHP-Rain. For the SPA-data, training set is cropped from 29,500 images and 1000 rainy images are used for testing. For the GT-Rain, 89 sequences are used for training and 7 sequences are used for testing. For the LHP-Rain, 2100 sequences are used for training and 300 sequences are used for testing. To evaluate the deraining performance on real scenes, we choose typical real rainy images from Internet-data. For single image deraining methods, we select the representative supervised deraining methods, including the CNN-based SPANet\cite{wang2019spatial}, PReNet\cite{ren2019progressive}, RCDNet\cite{wang2020model}, JORDER-E\cite{yang2019joint}, MPRNet\cite{mehri2021mprnet}, GT-Rain\cite{ba2022gt-rain}, transformer-based Uformer\cite{wang2022uformer} and IDT\cite{xiao2022image}. 

\noindent\textbf{Evaluation metrics.} We employ the full-reference PSNR and SSIM to evaluate the single image deraining results. Moreover, mean Average Precision (mAP) and Accuracy (Acc) are employed to evaluate object detection and lane segmentation after restoration by deraining methods.
\vspace{-0.1cm}
\subsection{Quantitative Evaluation}
\noindent\textbf{Deraining results on benchmarks.} We make comparisons with state-of-the-art deraining methods on three datasets SPA-data (A), GT-Rain (B) and LHP-Rain (C). The quantitative results are reported in Table \ref{tab2} under the following columns: A$\rightarrow$A, B$\rightarrow$B and C$\rightarrow$C. It is observed that transformer-based methods perform better than most CNN-based methods except for MPRNet in terms of PSNR and SSIM because of the superior representation of self-attention. Note that SCD-Former outperforms the existing state-of-the-art methods on all benchmarks, which confirms the effectiveness of our method with both self and cross-layer attention.
\vspace{-0.5cm}
\subsection{Qualitative Evaluation}
\noindent\textbf{Evaluation on LHP-Rain.} To further validate the deraining performance, we compare with the qualitative results of typical methods on LHP-Rain. As shown in Fig. \ref{RealRain}, SCD-Former achieves more visual pleasing results without rain residual and artifacts comparing with other methods, which cleans the rain streaks and veiling effect on the trees, highlight occlusion on the red car and the ground splashing water.

\noindent\textbf{Evaluation on real rainy images.} To evaluate the performance on real rainy images, we train SCD-Former on synthetic rain Rain100L\cite{yang2017deep}, real rain SPA-data, GT-Rain and LHP-Rain respectively and test on real rainy images. As shown in Fig. \ref{DatasetGeneralization}, the model trained on Rain100L performs poorly due to the huge domain gap. SPA-data and GT-Rain could remove real rain in the sky partially but they cannot handle the splashing water on the ground. The model trained on LHP-Rain has the best deraining performance which simultaneously removes rain streaks, veiling and ground splashing water without destroying image details.

  \begin{figure}[t]
   \vspace{0cm}  
\setlength{\abovecaptionskip}{0.1 cm}   
\setlength{\belowcaptionskip}{-0.2 cm}   
  \centering
     \includegraphics[width=1.0\linewidth]{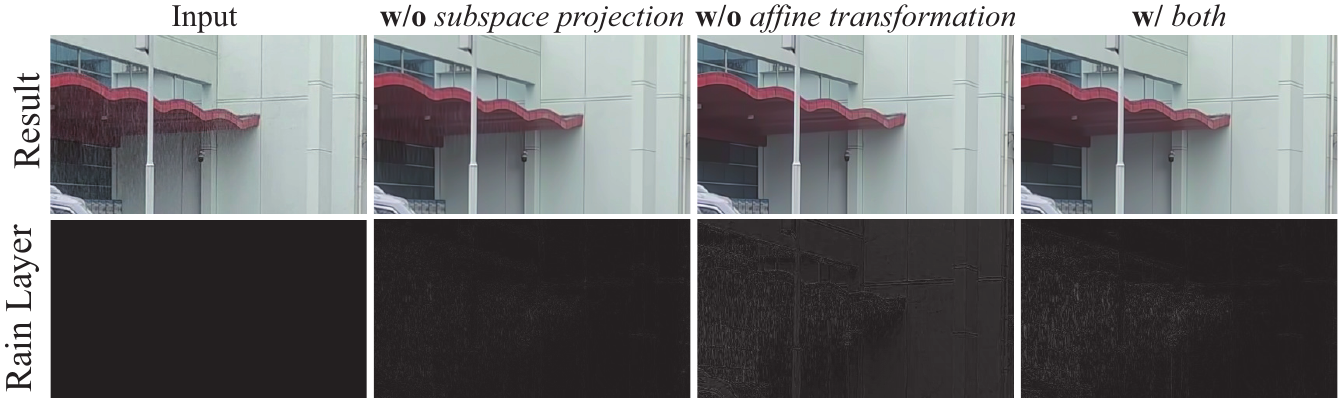}
  \caption{Ablation study of the robust low-rank tensor recovery model. The first row represents the deraining results, and the second row is corresponding rain. The first column is the original rainy frame and the remaining three columns represent the model without subspace projection, without affine transformation and full RLRTR. }
  \label{Ablationstudy_tensor}
\end{figure}

\subsection{Ablation Study}
\noindent\textbf{Effectiveness of subspace projection.} The subspace projection is used to characterize the extreme global low-rank property along the temporal dimension. In Fig. \ref{Ablationstudy_tensor}, there are obvious rain residual without subspace projection, implying that it is insufficient to characterize the property of the temporal dimension relying on the local prior of the temporal smoothness. Since the temporal low-rank property is neglected, leading to the rain residual in the results.

\noindent\textbf{Effectiveness of affine transformation.} The affine transformation is exerted to guarantee the pixel-level alignment of rainy video. As shown in Fig. \ref{Ablationstudy_tensor}, it can be observed that background residual and distortion obviously exist in the rain layer without affine transformation, because the extreme low-rank property along the temporal dimension is affected by data alignment among each frame in the video.
\begin{table}[t]
\small
  \centering
  \setlength{\abovecaptionskip}{0.2cm}
  \setlength{\belowcaptionskip}{-0.2cm}
  \caption{Ablation study of cross-layer attention.}
  \begin{threeparttable}
    \setlength{\tabcolsep}{1.5mm}{
  \begin{tabular}{cc}
  \hlinew{1.5pt}
Cross-layer attention & PSNR / SSIM\\
     \hlinew{1.5pt}
- &33.92\; 0.9384\\
 \checkmark &\textbf{34.33}\; \textbf{0.9403}\\
     \hlinew{1.5pt}
  \end{tabular}}
  \end{threeparttable}
  \label{tab5}
  \vspace{-0.6cm}
 \end{table}

\noindent\textbf{Effectiveness of cross-layer attention.} The design aims to find the correlations between rain layer and image layer. After iterations, the image layer sub-network obtains more attention on the rain features brought by cross-layer attention. In Table \ref{tab5}, We show that the complementary guidance from rain layer promotes the restoration of image layer. 
\setlength{\parskip}{0em}
\subsection{Discussion}
\noindent\textbf{Rain diversity of different datasets.} The rain diversity of dataset can be validated by the experiment of training models on one dataset and testing on the other unseen datasets among SPA-data (A), GT-Rain (B) and LHP-Rain (C). As shown in Table \ref{tab2}, on one hand, the best results of C$\rightarrow$A methods positively improve the performance of A, while all A$\rightarrow$C results performing worse than degraded result of C. On the other hand, the best results of C$\rightarrow$B have promotion while B$\rightarrow$C drop severely. Therefore, the outcome proves that our LHP-Rain contains more diverse rain categories than others, because they could not handle extreme challenging cases such as occlusion and ground splashing.
 \begin{figure}[t]
\setlength{\abovecaptionskip}{0.1 cm}   
\setlength{\belowcaptionskip}{-0.2 cm}   
  \centering
     \includegraphics[width=1.0\linewidth]{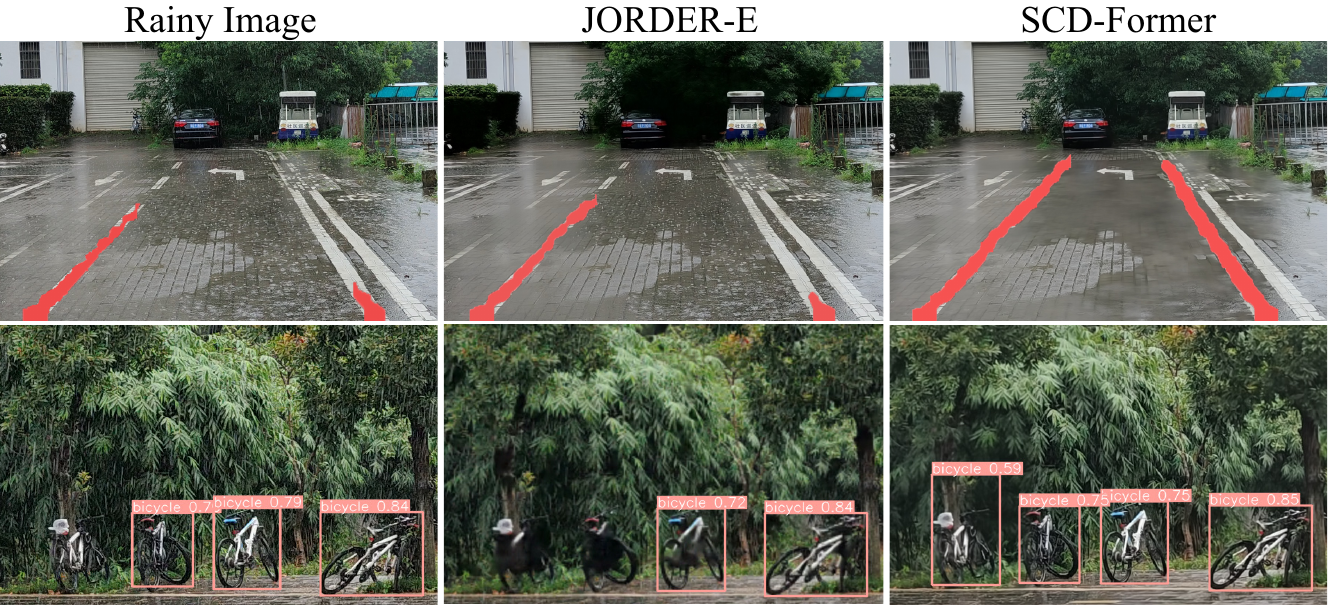}
  \caption{Evaluation of high-level tasks on lane segmentation and object detection. The performance improves significantly after removing rain streak and ground splashing from lanes and bicycles by our proposed SCD-Former.}
  \label{HighLevel}
\end{figure}

\begin{table}[t]
\footnotesize
  \centering
    \renewcommand\arraystretch{1.0}
  \setlength{\abovecaptionskip}{0pt}
  \setlength{\belowcaptionskip}{0pt}
  \caption{Evaluation of high-level tasks on deraining results.}
  \begin{threeparttable}
    \setlength{\tabcolsep}{1.0mm}{
  \begin{tabular}{ccccc}
   \hlinew{1.5pt}
Method& Det. (mAP)& Gain (Det.) & Seg. (Acc) & Gain (Seg.)\\
   \hlinew{1.5pt}
    Rainy&0.543&-&0.237&-\\
    SPANet&0.563&+0.020&0.268&+0.031\\
    PReNet&0.560&+0.022&0.255&+0.018\\
    RCDNet&0.556&+0.018&0.361&+0.124\\
    JORDER-E&0.568&+0.025&0.385&+0.148\\
    MPRNet&0.560&+0.017&0.350&+0.113\\
    Uformer-B&0.568&+0.025&0.306&+0.069\\
    IDT &0.570&+0.027&0.365&+0.128\\
    \textbf{SCD-Former}&\textbf{0.575}&\textbf{+0.031}&\textbf{0.449}&\textbf{+0.212}\\
       \hlinew{1.5pt}
  \end{tabular}}
  \end{threeparttable}
  \label{tab4}
  \vspace{-0.4cm}
 \end{table}
 
\noindent\textbf{Evaluation on downstream tasks.} We further evaluate the image deraining results on high-level tasks. For object detection, we apply the official YOLOv5 model on deraning results and report the mean average precision (mAP) of different classes in Table \ref{tab4}, where SCD-Former reaches the best average mAP among  typical objects. For lane segmentation, we choose the LaneNet\cite{neven2018towards} to predict the lane on LHP-Rain and SCD-Former has larger promotion on the segmentation accuracy. It is reasonable because SCD-Former performs well on removing ground splashing water and recovering the lane lines. The visualization results in Fig. \ref{HighLevel} shows that the lane on the surface of ground and the bicycles could be properly predicted after deraining by SCD-Former.

\noindent\textbf{User study on benchmarks quality.} We look for 126 volunteers to anonymously vote for the benchmark with best quality. Among existing benchmarks, we randomly select 100 samples and conduct user study including: rain diversity, image quality (resolution, JEPG, blur) and GT quality. The result is listed in Table \ref{userstudy}, where LHP-Rain consistently outperforms other benchmarks more than 50\%. 
 \begin{figure}[t]
 \vspace{-0.2cm}  
\setlength{\abovecaptionskip}{0.1 cm}   
\setlength{\belowcaptionskip}{-0.2 cm}   
  \centering
  \includegraphics[width=1.0\linewidth]{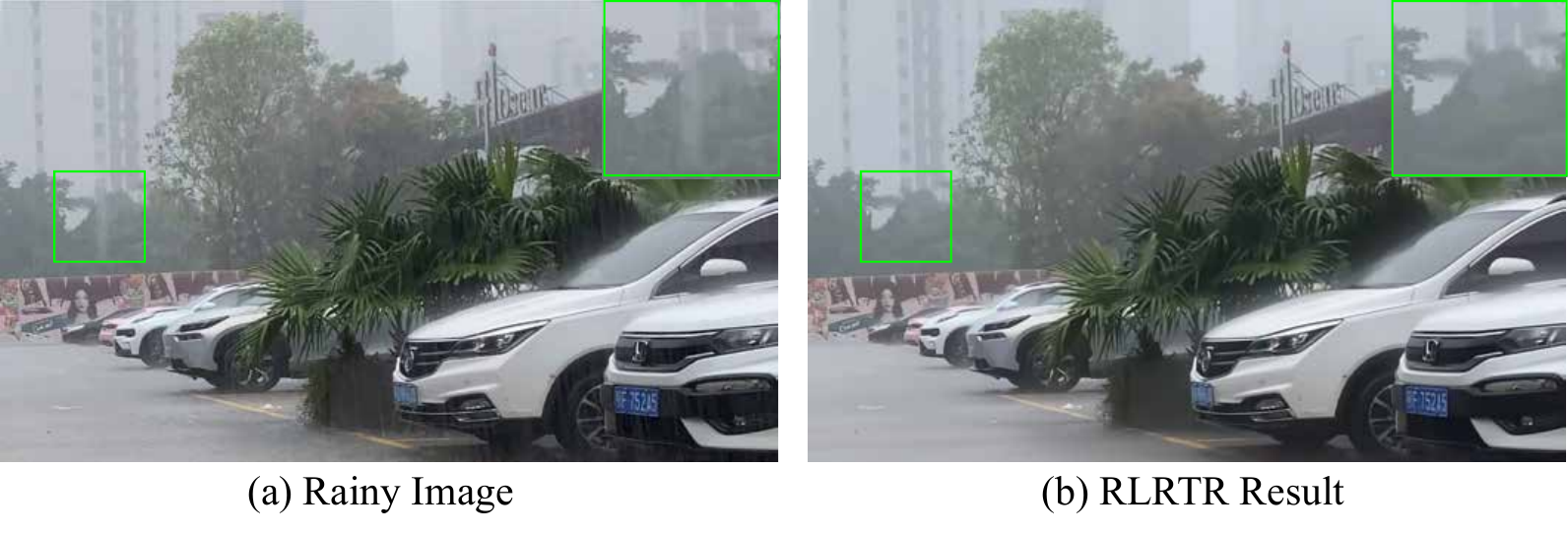}
   \caption{The limitation of RLRTR. The challenging occlusion effect could be removed by RLRTR from the background while the static haze is preserved.}
   \label{fig:limitation}
\end{figure}
\begin{table}[t]
\small
  \renewcommand\arraystretch{1.0}
  \centering
  \setlength{\abovecaptionskip}{0pt}
  \setlength{\belowcaptionskip}{0pt}
  \caption{User study on benchmarks quality.}
    \setlength{\tabcolsep}{0.8mm}{
  \begin{tabular}{ccccc}
 \hlinew{1.5pt}
 &\textbf{LHP-Rain}&SPA&RealRain1K&GT-Rain\\
 \hlinew{1.5pt}
Rain diversity&\textbf{63\%}&8\%&13\%&16\%\\
Image quality&\textbf{51\%}&14\%&20\%&15\%\\
GT quality&\textbf{55\%}&15\%&16\%&14\%\\
 \hlinew{1.5pt}
  \end{tabular}}
  \label{userstudy}
    \vspace{-0.5cm}
 \end{table}
  \vspace{-0.1cm}
 \section{Limitation}
 \vspace{-0.1cm}
Our proposed video deraining method is limited to remove the haze in the heavy rain scenes. RLRTR is adept at separating the static background from the dynamic rain. However, due to the steadiness of mist in the short interval, which is almost motionless in the background, RLRTR cannot decompose the haze from image layer well. Fig. \ref{fig:limitation} shows examples of rain and mist in our benchmark. Although the challenging rain patterns such as rain streaks are clearly removed, the result still contains haze. We look forward to handling the static haze in the future.    
 \vspace{-0.2cm}
\section{Conclusion}\label{sec7}
In this paper, we propose a large-scale and high-quality paired real rain benchmark. Our proposed LHP-Rain provides diverse rain categories, especially the ground splashing rain issue which is first claimed in deraining community. The model trained on LHP-Rain could generalize well on various real rainy scenes with great rain removal performance. Moreover, the proposed low-rank tensor recovery model could generate high-quality GT and detailed analysis confirms better results than others. In addition, we propose a single deraining baseline which performs well on removing rain from sky to the ground. Extensive experiments verify the superiority of the proposed benchmark and significantly improves segmentation task after removing splashing.

\noindent\textbf{Acknowledgements.} This work was supported in part by the National Natural Science Foundation of China under Grant 61971460 and Grant 62101294, in part by JCJQ Program under Grant 2021-JCJQ-JJ-0060 and in part by the Fundamental Research Funds for the Central Universities, HUST: 2022JYCXJJ001. 

{\small
\bibliographystyle{ieee_fullname}
\bibliography{egbib}
}

\end{document}